\documentclass[sigconf]{acmart}
\AtBeginDocument{%
  }

\setcopyright{acmlicensed}
\copyrightyear{2018}
\acmYear{2018}
\acmDOI{XXXXXXX.XXXXXXX}
\acmConference[Conference acronym 'XX]{Make sure to enter the correct
  conference title from your rights confirmation email}{June 03--05,
  2018}{Woodstock, NY}
\acmISBN{978-1-4503-XXXX-X/2018/06}

\usepackage{soul}
\usepackage{url}
\usepackage{graphicx}
\usepackage{subfigure}
\usepackage{amsthm}
\usepackage{makecell}
\usepackage{algorithm}
\usepackage{algorithmic}
\usepackage{color}
\usepackage{multirow}
\usepackage{natbib}
\usepackage{balance}

\begin{document}

\title{DARTer: Dynamic Adaptive Representation Tracker for Nighttime UAV Tracking}

\author{Xuzhao Li}
\authornote{Equal contribution.}
\authornote{Work as research intern in NTU.}
\affiliation{
  \institution{Nanyang Technological University}
  \city{Singapore}
  \country{Singapore}
}
\email{xuzhaoli2001@gmail.com}

\author{Xuchen Li}
\authornotemark[1]
\affiliation{
  \institution{Zhongguancun Academy}
  \city{Beijing}
  \country{China}
}
\email{s-lxc24@bjzgca.edu.cn}

\author{Shiyu Hu}
\authornote{Correspondence to Shiyu Hu.}
\affiliation{
  \institution{Nanyang Technological University}
  \city{Singapore}
  \country{Singapore}
}
\email{shiyu.hu@ntu.edu.sg}

\renewcommand{\shortauthors}{Xuzhao Li, Xuchen Li, and Shiyu Hu}

\begin{abstract}
Nighttime UAV tracking presents significant challenges due to extreme illumination variations and viewpoint changes, which severely degrade tracking performance. Existing approaches either rely on light enhancers with high computational costs or introduce redundant domain adaptation mechanisms, failing to fully utilize the dynamic features in varying perspectives. To address these issues, we propose \textbf{DARTer} (\textbf{D}ynamic \textbf{A}daptive \textbf{R}epresentation \textbf{T}racker), an end-to-end tracking framework designed for nighttime UAV scenarios. 
DARTer leverages a Dynamic Feature Blender (DFB) to effectively fuse multi-perspective nighttime features from static and dynamic templates, enhancing representation robustness. Meanwhile, a Dynamic Feature Activator (DFA) adaptively activates Vision Transformer layers based on extracted features, significantly improving efficiency by reducing redundant computations. Our model eliminates the need for complex multi-task loss functions, enabling a streamlined training process. 
Extensive experiments on multiple nighttime UAV tracking benchmarks demonstrate the superiority of DARTer over state-of-the-art trackers. These results confirm that DARTer effectively balances tracking accuracy and efficiency, making it a promising solution for real-world nighttime UAV tracking applications.
\end{abstract}



\maketitle

\section{Introduction}
\label{sec:introduction}
Unmanned aerial vehicle (UAV) tracking has widespread applications in aerial robotic vision, such as search and rescue \cite{al2019appearance} and traffic monitoring \cite{tian2011video}. With the advancement of deep learning \cite{he2016deep, alexey2020image} and large-scale datasets \cite{muller2018trackingnet, li2024dtllm, hu2024multi, li2024texts, huang2019got, li2024dtvlt}, daytime UAV trackers have achieved remarkable performance. However, nighttime UAV tracking remains a significant challenge due to extreme illumination variations, reduced contrast, and drastic viewpoint changes, which severely degrade tracking performance. State-of-the-art (SOTA) trackers \cite{Cao2022TCTrackTC, cao2023towards, cao2021siamapn++, feng2024memvlt} designed for daytime scenarios struggle to handle these conditions and often fail entirely. This underscores the urgent need for robust and efficient nighttime UAV tracking algorithms that can effectively enhance the applicability and survivability of UAVs in low-light environments.

Several approaches have been explored to address nighttime UAV tracking. One category involves light enhancement-based methods, which increase image brightness and subsequently apply daytime trackers. For example, \cite{fu2022highlightnet} employs a light enhancer to illuminate object areas, while \cite{li2021adtrack} integrates a low-light enhancer with correlation filtering-based tracking. Although these methods enable nighttime tracking, they heavily rely on additional enhancement networks, making end-to-end training challenging and increasing computational cost. Another category is domain adaptation-based methods, which aim to bridge the domain gap between day and night environments. TDA-Track \cite{fu2024prompt} incorporates temporal context information within a prompt-driven adaptation framework, while \cite{ye2022unsupervised} leverages domain adaptation techniques to refine nighttime object representations. However, these methods require large-scale, high-quality nighttime training data, which is often scarce and expensive to obtain. Despite these advancements, existing methods fail to fully utilize the dynamic feature variations across different viewpoints, which are crucial for improving tracking robustness. DCPT \cite{zhu2024dcpt} employs prompt-based learning to model nighttime tracking, but its reliance on dark clue prompts introduces significant computational redundancy. Similarly, \cite{wu2024mambanut} adopts adaptive curriculum learning to enhance tracking performance, but this increases optimization complexity and model overhead.

To overcome these limitations, we propose DARTer (Dynamic Adaptive Representation Tracker), an end-to-end nighttime UAV tracking framework that effectively captures dynamic multi-perspective features while maintaining computational efficiency. Specifically, DARTer employs a Dynamic Feature Blender (DFB) to fuse multi-view nighttime features from static and dynamic templates, enhancing feature representation robustness. Additionally, a Dynamic Feature Activator (DFA) adaptively activates Vision Transformer layers based on the extracted features, significantly improving efficiency by reducing redundant computations. Unlike previous methods that suffer from high training costs or excessive computational overhead, DARTer achieves a balanced trade-off between tracking accuracy and efficiency.

\begin{figure*}[htbp!]
  \centering
  \includegraphics[width=0.85\linewidth]{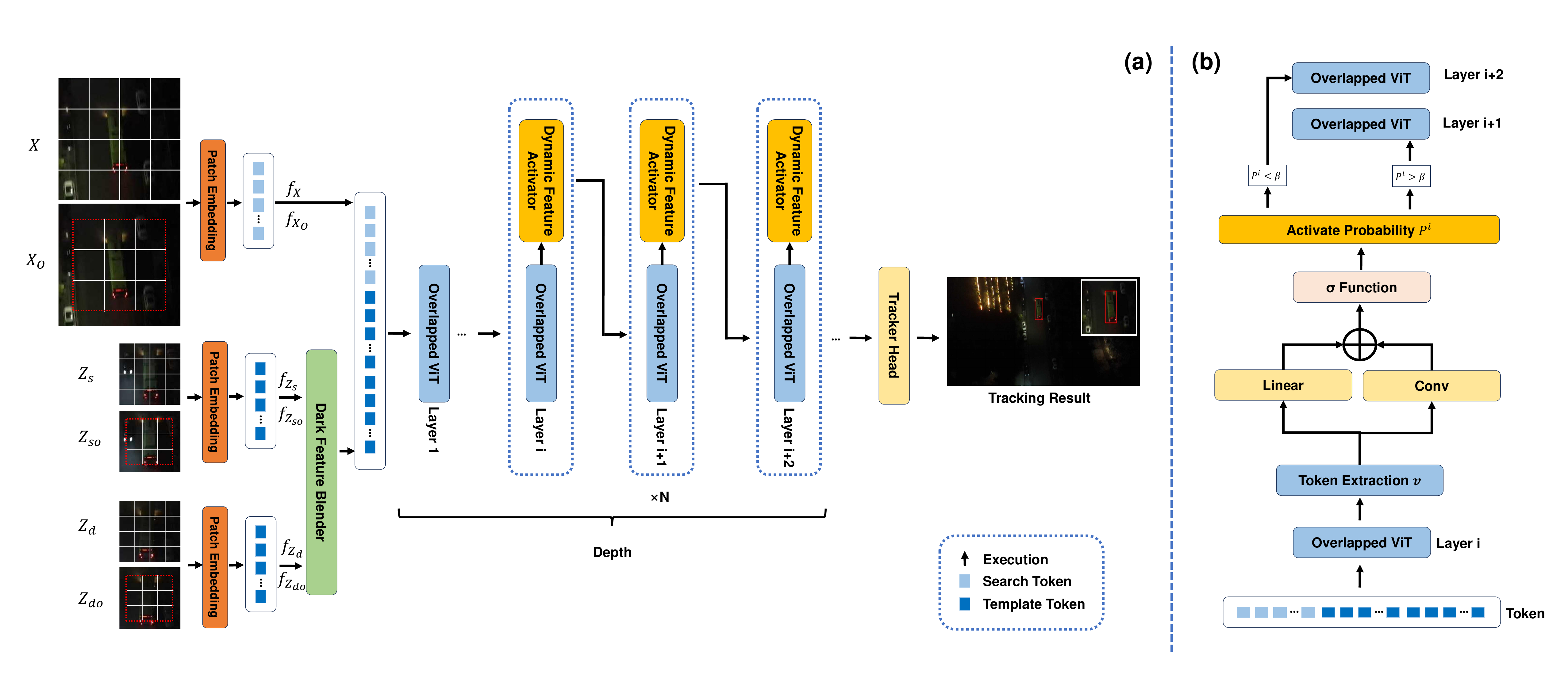}
  \caption{\textbf{(a) Overview architecture of DARTer.} The nighttime dynamic features of the static and dynamic templates are fused. The ViT blocks are dynamically activated according to the currently extracted nighttime features. \textbf{(b) Diagram of Dynamic Feature Activator.} The DFA module performs token extraction, transforms them through linear and convolution operations, and then conducts an activation process to adaptively select ViT layers and improve efficiency.}
   \label{fig:overview}
\end{figure*}

Extensive experiments on five nighttime UAV tracking benchmarks demonstrate that DARTer surpasses SOTA trackers, achieving a 6.3\% improvement in precision on NAT2021-L \cite{ye2022unsupervised}, showcasing its robustness in complex nighttime environments. These results confirm that DARTer provides a practical and effective solution for real-world nighttime UAV tracking applications.

\section{Methods}
\label{sec:methodology}
We propose a single-stream tracking framework named DARTer. Its architecture is illustrated in Fig. \ref{fig:overview}. The framework takes the search image, static and dynamic template images as inputs, and these images are sliced into overlapping patches. We use a Dark Feature Blender for static and dynamic templates to fuse and extract the nighttime features in different views, and then feed all images into Overlapped ViT \cite{shen2024overlapped} to extract dynamic features and match templates. Among them, We use the Dynamic Feature Activator to adaptively activate the ViT blocks and improve the efficiency of feature extraction. The details of these components will be described in the following subsections.

\subsection{Dark Feature Blender}
Before introducing the Dark Feature Blender (DFB), we introduce the input patches. The input images include the initial search image $X$, the static template $Z_s$ and the dynamic template $Z_d$. Meanwhile, these images are sliced into Overlapped patches \cite{shen2024overlapped}, i.e., O patches, including $X_o$, $Z_{so}$ and $Z_{do}$, respectively. These O patches connect the patches of the initial images and strengthen the associations among the image patches, making it easier to extract the dynamic features contained in different perspectives of the current static and dynamic templates.

To further learn and understand the state changes and essential characteristics of the object in different views, make full use of templates, and enhance the robustness of feature representation, we use a DFB module to perform feature fusion on the current static and dynamic templates. Specifically, we perform cross-attention operations on the features $f_{Z_s}$ and $f_{Z_d}$ corresponding to the initial static and dynamic templates, and obtain the nighttime fusion feature $f_Z$. The computational process of the initial static and dynamic templates is as follows:
\begin{equation}
\label{equ:fzs}
  f_{Z_{s'}} = \Phi_{CA}(f_{Z_s},f_{Z_d}),
\end{equation}
\begin{equation}
\label{equ:fzd}
  f_{Z_{d'}} = \Phi_{CA}(f_{Z_d},f_{Z_s}),
\end{equation}
\begin{equation}
\label{equ:dfb}
  f_Z = Concat(f_{Z_{s'}},f_{Z_{d'}}),
\end{equation}
where $\Phi_{CA}$ represents the cross-attention operation. In this operation, the first element functions as Q, and the second element is used to acquire K and V \cite{vaswani2017attention}. 
Similarly, after performing the same operations on the features $f_{Z_{so}}$ and $f_{Z_{do}}$ corresponding to the overlapped templates, we obtain nighttime fusion overlapped feature $f_{Z_{o}}$.

The dynamic template is updated at fixed intervals. This integrates the different perspectives and dynamic information of the tracking object at different times during the tracking process, enabling more comprehensive extraction of dynamic features during the feature fusion process, boosting the robustness of feature representation.

\begin{table*}[htbp!]
\centering
\caption{State-of-the-art comparison on the NAT2024-1 \cite{fu2024prompt}, NAT2021 \cite{ye2022unsupervised} and UAVDark135 \cite{li2022all} benchmarks. The top three results are highlighted in \textcolor{red}{\textbf{red}}, \textcolor{blue}{\textbf{blue}} and \textcolor{green}{\textbf{green}}, respectively. Note that the percent symbol (\%) is excluded for precision score (P), normalized precision (P\(_\text{Norm}\)) and area under the curve (AUC).}
\label{tab:overall_performance}
\resizebox{6in}{!}{
\begin{tabular}{c|c|ccc|ccc|ccc|cc} 
\toprule
\multirow{2}{*}{Tracker} & \multirow{2}{*}{Source} & \multicolumn{3}{c|}{NAT2024-1} & \multicolumn{3}{c|}{NAT2021} & \multicolumn{3}{c|}{UAVDark135} & \multirow{2}{*}{Avg.FPS} & \multirow{2}{*}{Params.(M)}  \\
& & P & P\(_{\text{Norm}}\) & AUC & P & P\(_{\text{Norm}}\) & AUC & P & P\(_{\text{Norm}}\) & AUC & &  \\
\hline
TCTrack \cite{Cao2022TCTrackTC} & CVPR 22 & 74.4 & 51.2 & 47 & 60.8 & 51.9 & 40.8 & 49.8 & 50.0 & 37.7 & 136 & 8.5 \\
TCTrack++ \cite{cao2023towards} & TPAMI 23 & 70.5 & 50.8 & 46.6 & 61.1 & 52.8 & 41.7 & 47.4 & 47.4 & 37.8 & 122 & 8.8 \\
MAT \cite{zhao2023representation} & CVPR 23 & 80.5 & \textcolor{green}{\textbf{76.3}} & 61.9 & 64.8 & 58.8 & 47.7 & 57.2 & 57.6 & 47.1 & 56 & 88.4 \\
HiT-Base \cite{Kang2023ExploringLH} & ICCV 23 & 62.7 & 56.9 & 48.2 & 49.3 & 44.2 & 36.4 & 48.9 & 48.7 & 41.1 & 156 & 42.1 \\
Aba-ViTrack \cite{li2023adaptive} & ICCV 23 & 78.4 & 72.2 & 60.1 & 60.4 & 57.3 & 46.9 & 61.3 & 63.5 & 52.1 & 134 & 7.9 \\
SGDViT \cite{yao2023sgdvit} & ICRA 23 & 53.1 & 47.2 & 38.1 & 53.1 & 47.9 & 37.5 & 40.2 & 40.6 & 32.7 & 93 & 23.3 \\
TDA-Track \cite{fu2024prompt} & IROS 24 & 75.5 & 53.3 & 51.4 & 61.7 & 53.5 & 42.3 & 49.5 & 49.9 & 36.9 & 114 & 9.2 \\
AVTrack-DeiT \cite{lilearningicml} & ICML 24 & 75.3 & 68.2 & 56.7 & 61.5 & 55.6 & 45.5 & 58.6 & 59.2 & 47.6 & 212 & 7.9 \\
DCPT \cite{zhu2024dcpt} & ICRA 24 & \textcolor{green}{\textbf{80.9}}  & 75.4 & \textcolor{green}{\textbf{62.1}} & \textcolor{green}{\textbf{69.0}} & \textcolor{green}{\textbf{63.5}} & \textcolor{blue}{\textbf{52.6}} & \textcolor{green}{\textbf{69.2}} & \textcolor{blue}{\textbf{69.8}} & \textcolor{green}{\textbf{56.7}} & 35 & 92.9 \\
MambaNUT \cite{wu2024mambanut} & arXiv 24 & \textcolor{blue}{\textbf{83.3}} & \textcolor{blue}{\textbf{76.9}} & \textcolor{blue}{\textbf{63.6}} & \textcolor{blue}{\textbf{70.1}} & \textcolor{red}{\textbf{64.6}} & \textcolor{green}{\textbf{52.4}} & \textcolor{blue}{\textbf{70.0}} & \textcolor{green}{\textbf{69.3}} & \textcolor{blue}{\textbf{57.1}} & 72 & 4.1   \\
\textbf{DARTer} & \textbf{Ours} & \textcolor{red}{\textbf{85.2}} & \textcolor{red}{\textbf{80.1}} & \textcolor{red}{\textbf{65.6}} & \textcolor{red}{\textbf{70.2}} & \textcolor{blue}{\textbf{63.7}} & \textcolor{red}{\textbf{53.2}} & \textcolor{red}{\textbf{71.6}} & \textcolor{red}{\textbf{72.1}} & \textcolor{red}{\textbf{58.2}} & 74 & 80.9 \\
\bottomrule
\end{tabular}}
\end{table*}

\subsection{Dynamic Feature Activator}
To fully extract and utilize nighttime dynamic features while enhancing the tracking efficiency, we propose a Dynamic Feature Activation (DFA) module, as shown in Fig. \ref{fig:overview} (b). This module calculates based on the dynamic fusion features of the previous ViT block to determine whether to activate the next ViT block. We feed all the search and template tokens into the DFA module, and obtain the activation probability. If the next ViT block is not activated, this block will be skipped directly.

Specifically, consider the \(i\)-th layer. Suppose that all the tokens of the output of the \((i - 1)\)-th layer are denoted as $t_{1:k}^{i-1}(f_D)$, where \(k\) is the number of tokens, and $f_D = Concat(f_X,f_{X_o},f_Z, f_{Z_o})$. Define a feature extraction vector \(v\) belonging to the standard normal distribution \(N(0, 1)\). Then the input of the \(i\)-th layer is \(r^{i} = vt_{1:k}^{i-1}(f_D)\), and the activation probability \(p^i\) of the \(i\)-th layer ViT block is as follows:
\begin{equation}
\label{equ:p}
  p^i=\sigma(L(r^{i}) + Conv(r^{i})),
\end{equation}
where \(\sigma\) represents \(\frac{1}{2}(\tanh + 1)\), $L$ represents the linear operation, $Conv$ represents the convolution operation, and the activation probability is \(p^i\in(0, 1)\). Let \(\beta\) be the activation threshold. If \(p^i>\beta\), then the \(i\)-th layer is activated; otherwise, the output of the \((i - 1)\)-th layer is directly fed into the \((i + 1)\)-th layer, and the activation judgment is carried out again.

The initial ViT blocks extract the basic features of the image, which play a crucial role in subsequent template matching. To avoid the situation where all blocks are not activated, we perform feature activation calculations on all blocks except the first ViT block. 

\subsection{Prediction Head and Training Loss}
Similar to the corner detection head \cite{cui2022mixformer, ye2022joint}, we use a bounding-box prediction head \(H\) with four stacked Conv-BN-ReLU layers. First, we convert the output tokens of the search image into a 2D spatial feature map. Inputting these features into the prediction head, we get a local offset $o$, a normalized bounding-box size $s$, and an object classification score $p$ as the prediction results. We estimate the object by finding the location with the highest classification score.

Regarding the training loss, DARTer combines the softmax cross-entropy loss \cite{wei2023autoregressive} and the SloU loss \cite{gevorgyan2022siou}. The loss function for the training phase is $L_{\text{total}}=\lambda_{1}L_{ce}+\lambda_{2}L_{SloU}$, where $\lambda_1$ and $\lambda_2$ are the weights assigned to the two losses. In our experiments, we set $\lambda_1 = 2$ and $\lambda_2 = 2$. Obviously, there is no need for us to rely on complex hand-designed loss functions.

\section{Experiment}
\label{sec:experiment}

\subsection{Implementation Details}
We use Overlapped ViT \cite{shen2024overlapped} as the backbone. The activation probability threshold $\beta = 0.3$. The image sizes of the search and template are $128 \times 128$ and $256 \times 256$, respectively. The patch size is $16 \times 16$. The initial and O patches of the search image are $16 \times 16$ and $15 \times 15$, and the initial and O patches of the template are $8 \times 8$ and $7 \times 7$, respectively. We use four common datasets and three nighttime datasets for training, including LaSOT \cite{fan2019lasot}, GOT10K \cite{huang2019got}, COCO \cite{lin2014microsoft}, TrackingNet \cite{muller2018trackingnet} and BDD100K\_Night \cite{yu2020bdd100k}, SHIFT\_night \cite{sun2022shift}, ExDark \cite{loh2019getting}. The model is trained for 150 epochs using the AdamW optimizer \cite{loshchilov2017decoupled}, with a batch size of 32. Each epoch involves 60,000 sampling pairs. The initial learning rate is set to 0.0001, and after 120 epochs, the learning rate decays at a rate of 10\%. The model is trained on a server with four A5000 GPUs and tested on an RTX-3090 GPU.

\subsection{Comparison Results}
We evaluate DARTer on five benchmarks, including NAT2024-1 \cite{fu2024prompt}, NAT2021 \cite{ye2022unsupervised}, UAVDark135 \cite{li2022all}, NAT2021-L \cite{ye2022unsupervised} and DarkTrack2021 \cite{ye2022tracker}. We then compare DARTer with the current state-of-the-art (SOTA) trackers.

\textbf{NAT2024-1.}  NAT2024-1 \cite{fu2024prompt} is a large-scale, long-duration nighttime UAV tracking benchmark. This benchmark has been meticulously designed to comprehensively evaluate the performance of tracking algorithms. As presented in Tab. \ref{tab:overall_performance}, our DFTrack outperforms the other SOTA trackers in this benchmark. Specifically, it has a precision score (P) of 85.2\%, a normalized precision (P\(_\text{Norm}\)) of 80.1\% and an area under the curve (AUC) of 65.6\%. DFTrack surpasses the SOTA tracker by 1.9\%, 3.2\% and 2.0\%, respectively. This result clearly demonstrates the effectiveness of the methods we proposed.  

\textbf{NAT2021 and NAT2021-L.} NAT2021 \cite{ye2022unsupervised} and NAT2021-L \cite{ye2022unsupervised} are typical nighttime UAV tracking benchmarks with diverse image attributes, like high occlusion and complex environmental elements. Despite the challenges, our tracker has achieved remarkable results. Among all the trackers, the AUC score is 53.2\%, achieving the best performance in NAT2021. As shown in Tab. \ref{tab:nat2021l_performance}, DARTer has demonstrated surprising results in NAT2021-L. It ranks first in P and P\(_\text{Norm}\) and AUC, outperforming the previous SOTA model.

\begin{table}[htbp!]
\centering
\caption{Comparison on the NAT2021-L \cite{ye2022unsupervised} benchmark. The top three results are highlighted in \textcolor{red}{\textbf{red}}, \textcolor{blue}{\textbf{blue}} and \textcolor{green}{\textbf{green}}, respectively.}
\label{tab:nat2021l_performance}
\resizebox{2.4in}{!}{
\begin{tabular}{c|c|cccc} 
\toprule
\multirow{2}{*}{Tracker} & \multirow{2}{*}{Source} & \multicolumn{3}{c}{NAT2021-L} \\
& & P & P\(_{\text{Norm}}\) & AUC\\
\hline
SiamRPN++ \cite{li2019siamrpn++} & CVPR 19 & 42.9 & 35.8 & 30.0 \\
Ocean \cite{zhang2020ocean} & ECCV 20 & 45.1 & 40.0 & 31.6 \\
HiFT \cite{cao2021hift} & ICCV 21 & 43.0 & 33.0 & 28.8 \\
SiamAPN \cite{fu2021siamese} & ICRA 21 & 37.7 & 27.7 & 24.2 \\
SiamAPN++ \cite{cao2021siamapn++} & IROS 21 & 40.0 & 32.7 & 28.0 \\
UDAT-BAN \cite{ye2022unsupervised} & CVPR 22 & 49.4 & 43.7 & 35.3 \\
UDAT-CAR \cite{ye2022unsupervised} & CVPR 22 & \textcolor{green}{\textbf{50.4}} & \textcolor{green}{\textbf{44.7}} & \textcolor{green}{\textbf{37.8}} \\
DCPT \cite{zhu2024dcpt} & ICRA 24  & \textcolor{blue}{\textbf{58.6}} & \textcolor{blue}{\textbf{54.6}} & \textcolor{blue}{\textbf{47.4}} \\
\textbf{DARTer} & \textbf{Ours} & \textcolor{red}{\textbf{64.9}} & \textcolor{red}{\textbf{58.6}} & \textcolor{red}{\textbf{50.9}} \\
\bottomrule
\end{tabular}}
\end{table}

\textbf{UAVDark135.} UAVDark135 \cite{li2022all} is widely used as a benchmark for nighttime tracking. As shown in Tab. \ref{tab:overall_performance}, the method we proposed outperforms other SOTA trackers. The P, P\(_\text{Norm}\) and AUC reach 71.6\%, 72.1\% and 58.2\%, respectively. We can see that DARTer can track objects in nighttime scenes more accurately.

\textbf{DarkTrack2021.} DarkTrack2021 \cite{ye2022tracker} is a highly challenging nighttime UAV tracking benchmark with many situations of interference. Nevertheless, as demonstrated in Tab. \ref{tab:darktrack2021_performance}, our model still shows outstanding performance. It reaches the SOTA level in  P\(_\text{Norm}\) and AUC. This indicates that the model we proposed has strong adaptability and robustness.

\begin{table}[htbp!]
\centering
\caption{Comparison on the DarkTrack2021 \cite{ye2022tracker} benchmark. The top three results are highlighted in \textcolor{red}{\textbf{red}}, \textcolor{blue}{\textbf{blue}} and \textcolor{green}{\textbf{green}}, respectively.}
\label{tab:darktrack2021_performance}
\resizebox{2.6in}{!}{
\begin{tabular}{c|c|cccc}  
\toprule
\multirow{2}{*}{Tracker} & \multirow{2}{*}{Source} & \multicolumn{3}{c}{NAT2021-L} \\
& & P & P\(_{\text{Norm}}\) & AUC\\
\hline
SiamRPN \cite{li2018high} & CVPR 18 & 50.9 & 48.5 & 38.7 \\
DIMP18 \cite{bhat2019learning} & ICCV 19 & 62.0 & 58.9 & 47.1 \\
PRDIMP50 \cite{danelljan2020probabilistic} & CVPR 20& 58.0 & 55.9 & 46.4 \\
SiamAPN++ \cite{cao2021siamapn++} & IROS 21 & 48.9 & 46.1 & 37.7 \\
HiFT \cite{cao2021hift} & ICCV 21  & 50.3 & 47.1 & 37.4 \\
SiamAPN++-SCT \cite{ye2022tracker} & RAL 22 & 53.7 & 51.1 & 40.8 \\
DIMP50-SCT \cite{ye2022tracker} & RAL 22 & \textcolor{red}{\textbf{67.7}} & \textcolor{green}{\textbf{63.3}} & \textcolor{green}{\textbf{52.1}} \\
DCPT \cite{zhu2024dcpt} & ICRA 24 & \textcolor{green}{\textbf{66.7}} & \textcolor{blue}{\textbf{64.6}} & \textcolor{blue}{\textbf{54.0}} \\
\textbf{DARTer} & \textbf{Ours} & \textcolor{blue}{\textbf{67.6}} & \textcolor{red}{\textbf{64.8}} & \textcolor{red}{\textbf{54.5}} \\
\bottomrule
\end{tabular}}
\end{table}

As demonstrated in Tab. \ref{tab:overall_performance}, our DARTer can run in real-time at over 74fps. Furthermore, the Precision of DARTer on NAT2024-1 \cite{fu2024prompt} is 1.9 \% higher than that of MambaNUT \cite{wu2024mambanut}. This demonstrates that our method can effectively utilize dynamic features and improve tracking efficiency and performance.

\begin{figure}[htbp!]
  \centering   
  \includegraphics[width=0.87\linewidth]{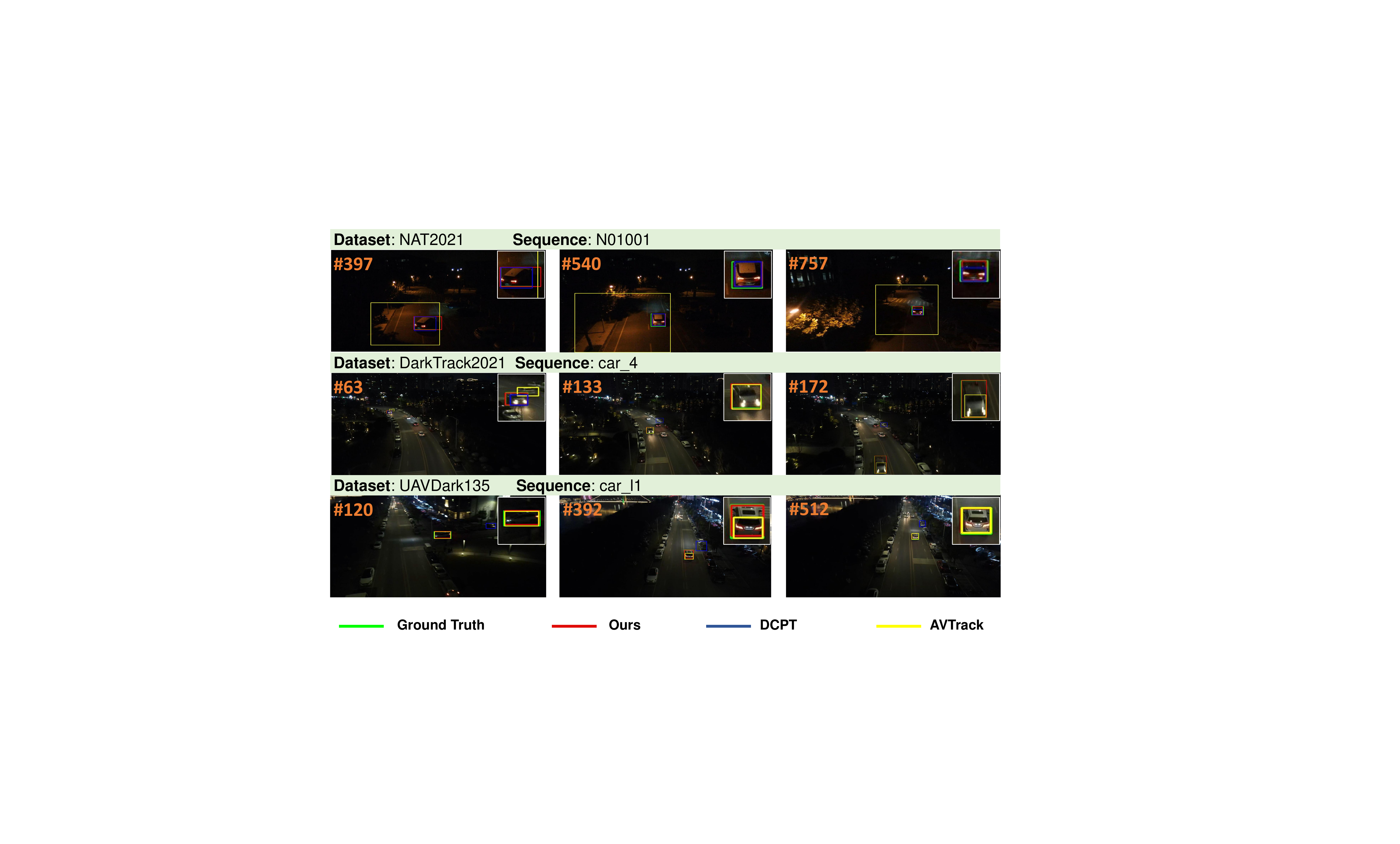}
   \caption{Qualitative comparison results of our tracker with other two latest trackers (i.e., DCPT \cite{zhu2024dcpt} and AVTrack \cite{lilearningicml} in representative nighttime scenarios. Better viewed in color with zoom-in.}
   \label{fig:vis}
   \vspace{-10pt}
\end{figure}

As depicted in Fig. \ref{fig:vis}, we also visualize the tracking results of our model and the two previous SOTA models on three representative nighttime scenarios from NAT2021 \cite{ye2022unsupervised}, DarkTrack2021 \cite{ye2022tracker} and UAVDark135 \cite{li2022all}. These sequences have small, distant objects captured by UAVs at night, with interference from similar objects. Clearly, our model has higher tracking accuracy and stronger robustness, proving the effectiveness of our proposed modules in night tracking.

\subsection{Ablation Study and Analysis}
\label{sec:ablation}
The Dark Feature Blender (DFB) and Dynamic Feature Activator (DFA) modules serve as the core components of our tracker. The DFB fully leverages dynamic features from different views. Meanwhile, it enhances the extraction and learning of nighttime features, boosting the robustness of feature representation. As shown in Tab. \ref{tab:ablation_module}, the DFB enhances the basic tracker, increasing the success score on NAT2024-1 by 1.95\%. The DFA, via its adaptive activation mechanism, improves the template matching efficiency. It also ensures the perception of nighttime objects and enhances the AUC, P\(_{\text{Norm}}\) and P. Ultimately, the performance of the model has been significantly improved compared to the baseline.

\begin{table}[htbp!]
\centering
\caption{Impact of DFB and DFA on the performance of the baseline trackers on NAT2024-1.}
\vspace{-5pt}
\resizebox{2.6in}{!}{
\begin{tabular}{l c c c c c}
\toprule
Method & DFB & DFA & P & P\(_{\text{Norm}}\) & AUC\\
\midrule   
\multirow{4}{*}{DARTer} & $\checkmark$ & $\checkmark$ & $85.2$ & $80.1$ & $65.6$ \\
&  & $\checkmark$ & $84.3_{\downarrow0.9}$ & $79.5_{\downarrow0.6}$ & $64.6_{\downarrow1.0}$ \\
& $\checkmark$ &  & $83.4_{\downarrow1.8}$ & $78.6_{\downarrow1.5}$ & $64.2_{\downarrow1.4}$ \\
&  &  & $81.5_{\downarrow3.7}$ & $76.9_{\downarrow3.2}$ & $62.3_{\downarrow3.3}$ \\
\bottomrule
\end{tabular}}
\vspace{-5pt}
\label{tab:ablation_module}
\end{table}

\section{Conclusion}
\label{sec:conclusion}
We propose DARTer (Dynamic Adaptive Representation Tracker), an end-to-end framework for nighttime UAV tracking that integrates the Dynamic Feature Blender (DFB) for multi-perspective feature fusion and the Dynamic Feature Activator (DFA) for adaptive Vision Transformer activation, enhancing feature robustness while reducing computational redundancy. Extensive experiments on five major nighttime UAV tracking benchmarks demonstrate that DARTer achieves state-of-the-art performance, confirming its effectiveness in balancing tracking accuracy and efficiency. By advancing feature fusion and adaptive computation in nighttime tracking, DARTer contributes to the broader field of low-light visual perception and efficient transformer-based tracking. We believe this work will inspire further research in adaptive feature modeling, lightweight transformer architectures, and robust tracking under extreme conditions, fostering new developments in real-world UAV applications and beyond.
\clearpage

\bibliographystyle{ACM-Reference-Format}
\balance
\bibliography{sample-base}

\end{document}